\documentclass[conference]{IEEEtran}
\IEEEoverridecommandlockouts 

\ifCLASSINFOpdf
\else
   \usepackage[dvips]{graphicx}
\fi
\usepackage{url}

\usepackage{graphicx} 
\usepackage{amsmath}
\usepackage[top=0.756in, bottom=1.033in, right=0.635in, left=0.635in, paper=letterpaper]{geometry} 
\usepackage{algorithm}
\usepackage[noend]{algpseudocode}
\usepackage{subcaption}
\usepackage{ amssymb }
\usepackage{amsmath}
\usepackage{hyperref}
\usepackage{dsfont}
\usepackage{xcolor}
\usepackage{setspace}
\usepackage{lipsum}
\usepackage{float}
\usepackage{geometry} 

\makeatletter
\def\BState{\State\hskip-\ALG@thistlm}
\makeatother

\begin{document}

\title{Smart Information Exchange for Unsupervised Federated Learning via Reinforcement Learning
\thanks{This project was supported in part by NSF under grant CNS-2212565 and ONR under grant N000142112472.}
}
\author{\IEEEauthorblockN{Seohyun Lee\IEEEauthorrefmark{1}, Anindya Bijoy Das\IEEEauthorrefmark{2}, Satyavrat Wagle\IEEEauthorrefmark{2}, Christopher G. Brinton\IEEEauthorrefmark{2}} 
\IEEEauthorblockA{\IEEEauthorrefmark{1}Department of Computer Science, Purdue University \\
\IEEEauthorrefmark{2}Elmore Family School of Electrical and Computer Engineering, Purdue University\\
\{\texttt{lee3296, das207, wagles, cgb}\}@purdue.edu
}}

\IEEEtitleabstractindextext{%

\begin{abstract}
One of the main challenges of decentralized machine learning paradigms such as Federated Learning (FL) is the presence of local non-i.i.d. datasets. Device-to-device transfers (D2D) between distributed devices has been shown to be an effective tool for dealing with this problem and robust to stragglers. In an unsupervised case, however, it is not obvious how data exchanges should take place due to the absence of labels. In this paper, we propose an approach to create an optimal graph for data transfer using Reinforcement Learning. The goal is to form links that will provide the most benefit considering the environment's constraints and improve convergence speed in an unsupervised FL environment. Numerical analysis shows the advantages in terms of convergence speed and straggler resilience of the proposed method to different available FL schemes and benchmark datasets.
\end{abstract}
\vspace{0.02in}

}

\maketitle
\IEEEdisplaynontitleabstractindextext
\IEEEpeerreviewmaketitle

\section{Introduction}
Federated Learning (FL) \cite{1} has emerged as a popular paradigm for training machine learning models across decentralized devices while preserving local data privacy. In FL, participating devices collaboratively build a global model by exchanging model updates. This approach circumvents the need for centralized data aggregation, which is valuable when data cannot be easily transferred due to privacy concerns, bandwidth limitations, or regulatory constraints.

One of the challenges in FL is the presence of non-i.i.d. data distributions across devices, which can slow convergence speed and result in global model bias \cite{2}. A recent trend of work has considered mitigating these issues through augmenting FL with device-to-device (D2D) communications in relevant network settings, e.g., wireless sensor networks \cite{3}. In D2D-enabled FL, information exchange is employed to reduce overfitting local datasets \cite{4}. However, factors such as knowing which data-points to exchange and the overlying communication costs from the communication between local devices have an impact on its efficacy. This is exacerbated with unlabeled data, where the ground-truths are unknown. 
Thus, it requires further investigation on how to facilitate discovery of an effective D2D structure in an unsupervised environment which would also lead to an improved convergence of the overall learning task.

Several studies have investigated communication-efficient strategies for FL to reduce the burden of transmitting large model updates over limited communication channels. For example, according to the approach in \cite{1}, client devices update a global model based on local updates. Additionally, other methods have focused on quantization \cite{6} and sparsification \cite{7} techniques to reduce the communication overhead. There has also been work in graph discovery with Reinforcement Learning (RL) \cite{5} for D2D exchanges in the supervised setting \cite{8}, and unsupervised D2D exchange over a randomized graph with contrastive learning (CL) \cite{9}. However, while CL is shown to be effective for learning discriminative representations, it does not explicitly try to reconstruct the input data. This can be a disadvantage for tasks where reconstruction is important. In this regard, to our best knowledge, no work has focused on graph discovery for D2D in an unsupervised FL setting with autoencoders, which is useful for tasks such as denoising \cite{10} or anomaly detection \cite{11}. This can be used to learn unsupervised representations of the data, which is useful for downstream tasks such as classification, clustering, and retrieval.

In this work, we combine RL with FL in a unique manner to reduce the communication burden. 
By introducing a diversity-driven metric that quantifies the dissimilarity among local datasets, we encourage the exchange of data-points between devices with distinct data characteristics. We emphasize on improving the convergence of the overall model, thereby enhancing the robustness and generalization of the global model.

\vspace{-0.15 cm}
\subsection{Summary of Contributions} 
\vspace{-0.15 cm}
\begin{itemize}
    \item In our proposed approach, we leverage Principal Component Analysis (PCA) to retain important features of local datasets and utilize $K$-means++ to characterize dataset dissimilarity between devices.
    
    \item We propose a decentralized RL methodology to discover an optimal graph for D2D exchange considering factors of local dataset dissimilarity and communication. In our scheme, devices act as RL agents, training locally for link formation and engaging in data and reward sharing. After D2D exchanges have taken place, we use autoencoders as our chosen unsupervised learning method. 
    
    \item We evaluate our method on the FashionMNIST and CIFAR-10 datasets. Our method shows improvement over baselines in terms of convergence speed with respect to the reconstruction loss and downstream classification tasks, and also provides robustness against the stragglers.
\end{itemize}
\newtheorem{assumption}{\textbf{Assumption}}

\vspace{-0.07 cm}
\section{System Model}

\subsection{Federated Learning Setup}

\begin{figure} 
    \centering
    \includegraphics[width=0.65\linewidth]{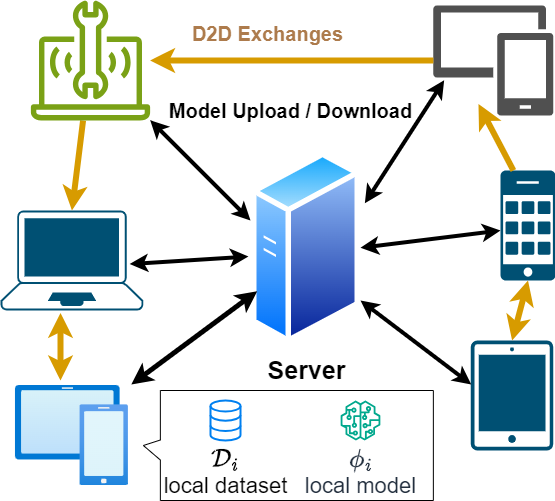}
    \caption{\small Environment for D2D-enabled FL for heterogeneous networks (a possible network may include phones, tablets or other devices). Each local device acts as its own agent in the RL formulation.}
    \vspace{-0.15 in} 
\end{figure}

We make our FL environment system a network of clients $C = \{c_1, c_2, ..., c_N\}$, local datasets $\mathcal{D}_i \in \mathcal{D}$ (where $\mathcal{D}$ denotes the full dataset), and local models $\phi_{i}^{\tau} \in \mathbb{R}^p$, where $p$ is the number of parameters in each model and $\tau$ is the time-step of training.
Local models $\{\phi_{i}^{\tau}\}_{c_i \in C}$ are aggregated every $\tau_a$ mini-batch iterations at server $\mathcal{S}$ to obtain a global model $\phi_{G}^{\tau}$, which is broadcasted back to all devices $c_i \in C$ for further local training. In this unsupervised setting, we have the following assumptions which will assist the overall framework to find an optimal D2D communication (Sec. \ref{sec:d2d}) while maintaining the security constraints (Sec. \ref{sec:sec}). Now we discuss characteristics of our framework that hold relevance in a FL task.

\begin{assumption}[\textbf{Sufficient Datapoints}]
\label{as:sufficientdata} 
Every local device $c_i$ (where $i = 1, 2, \dots, N$) has sufficient datapoints of all of its local clusters, such that engaging in data transfer will not affect the overall diversity--which we define as the number of clusters with more datapoints than a set threshold--of any client.
\end{assumption}

\begin{assumption}[\textbf{Number of Local Clusters}]
\label{as:clusters} 
Each client $c_i$ knows the number of clusters that best separate out the datapoints of its initial local dataset. \footnote{Methods such as the elbow method \cite{20}, where the within-cluster-sum-of-square (WCSS) forms an elbow hinge compared to other set cluster values, are commonly used to find the number of local clusters. Our graph discovery method is independent of the method for finding the number of clusters, and hence, in this work, we do not explore such methods in details.}
\end{assumption}

\subsection{Trust Between Devices}
\label{sec:sec}
In D2D communication, due to user privacy concerns, data privacy is integral to the overall framework. Certain data points cannot be transmitted unless permitted by the transmitter device. In this work, it is enforced through a local trust matrix which is denoted by $\mathbf{T}_j \in \mathbb{R}^{N \times k_j}$, where $k_j$ is the number of clusters in the local dataset, where $j = 1, 2, \dots, N$. The entries of trust matrix $\mathbf{T}_j$ belong to the set $\{0, 1\}$,
\begin{equation}
    \mathbf{T}_j[i, n] =
    \begin{cases}
    1 & \texttt{if } c_j \text{ trusts } c_i \text{ with cluster  } n \\
    0 & \texttt{otherwise}
\end{cases}.
\end{equation}
Thus, if $\mathbf{T}_j[i, n]=0$, it means transmitter device $c_j$ does not trust $c_i$ with its cluster $n$, where $n = 1, 2, \dots, k_j$. 

\subsection{D2D communication}
\label{sec:d2d}

We assume that D2D communication can be established
among the devices for datapoint exchange before the learning task. \cite{15} shows that small amount of datapoint exchange can reduce the non-i.i.d. skew in a decentralized learning environment. 
Similar to the approach in \cite{8}, we find the probability of unsuccessful transmission as 
$\mathbf{P}_D(i,j) = 1 - \text{exp}\left(\frac{-(2^r-1) \cdot \sigma^2}{\mathbf{W}_{ij}}\right)$, 
where $\textbf{W} \in \mathbb{R}^{N \times N}$, such that 
$\textbf{W}_{ij}$ defines the received signal strength (RSS) at $c_i$ when receiving a signal from $c_j$. 
For simplicity, we assume the same noise power of $\sigma^2$ across all channels and a constant rate of transmission $r$ between any two devices. We use this value for reward formulation in Sec \ref{sec:reward} and now discuss our RL setup for graph discovery.

\section{Optimal Graph Discovery}
\label{sec:reward}

In our RL setup, we define $M$ as the number of episodes we want an agent to experience before updating, $t \in [0, T]$ as the update interval of experience buffer $\textbf{B}_i \in \mathbb{R}^{M \times 2}$, which we define in Sec. \ref{sec:linkform}, and $e \in [0, E]$ as the current episode of the RL training process. State $s_i^{e} = \{\textbf{W}_{ij}\}_{j \in C} \in \mathbb{R}^N$ at device $c_i$ is defined as the RSS from $c_j$ and action $a_j^{e} \in \mathbb{R}^N$ is the incoming edge $j \rightarrow i$ predicted for episode $e$.

For the network graph, we define graph $\mathcal{G}$ such that the nodes, which are local devices $c_i \in C$, are connected if they have a direct communication link in set of edges $\mathcal{E}$. To achieve an optimal graph for data transfer, we define a reward function that encourages finding devices with markedly distinct data characteristics and reduces communication costs. A common way of dealing with unlabeled data is with clustering methods. However, when dealing with high-dimensional data, the sheer number of features can lead to potentially obscure patterns. In this aspect, Principal Component Analysis (PCA) \cite{16} can offer a solution by transforming the original feature space into a new space having reduced dimensionality while retaining the most significant information. We leverage the advantages of PCA to retain the important features and apply clustering using $K$-means++ \cite{17} to get the desired clusters for reward formulation.

Now assume that client $c_i$ has $k_i$ clusters, where we denote those corresponding centroid vectors as $\mathbf{v}_{in}$, where $n = 1, 2, \dots, k_i$. Note that client $c_i$ wants to receive datapoints of one or more such clusters from client $c_j$ which may not be present in the initial local dataset of $c_i$. We choose any cluster of client $c_j$ having centroid $\mathbf{v}_{jm}$, and define the distance threshold $\beta$ so that $\lambda_{ij_m}$ clusters in client $c_i$ are significantly different than that cluster. Thus, $\lambda_{ij_m} = \sum_{n = 1}^{k_i} \left( \mathds{1}_{||\mathbf{v}_{in} - \mathbf{v}_{jm}|| > \beta} \right)$. Now, we denote the number of clusters that $c_i$ can receive datapoints from $c_j$ as $\lambda_{ij}$. Depending on $\mathbf{T}_j [i,m]$, we can write
$\lambda_{ij} = \sum_{m = 1}^{k_j} \left( \mathds{1}_{\lambda_{ij_m} = k_i}  \times \mathbf{T}_j [i,m]\right)$, 
where $\lambda_{ij}$ is the number of clusters that the receiver device may be interested in receiving datapoints from. Then we use $\lambda_{ij}$ for our local reward formulation as
\begin{equation}
    r_{ij} = \alpha_1 \cdot \lambda_{ij} - \alpha_2 \cdot \mathbf{P}_D(i,j),\label{eq:3}
\end{equation}
where $\mathbf{P}_D(i,j)$ (obtained from Sec. \ref{sec:d2d}) is the probability of unsuccessful transmission between the local devices $i$ and $j$. To incorporate the above metrics, the reward function has user parameterized weights $\alpha_1, \alpha_2$ to consider the distance metric and failed transmission, which now we use to define our global reward.
\subsection{Deciding the Best Link}
\label{sec:linkform}
\begin{assumption}[\textbf{Incoming Edges}]
\label{as:incoming} 
The data diversity at each
receiver is impacted most by the first edge, after which the
change in diversity diminishes as the number of edges increase. Hence, each client $c_i$ is limited to one incoming edge.
\end{assumption}
\begin{figure} 
    \centering
    \includegraphics[width=0.92\linewidth]{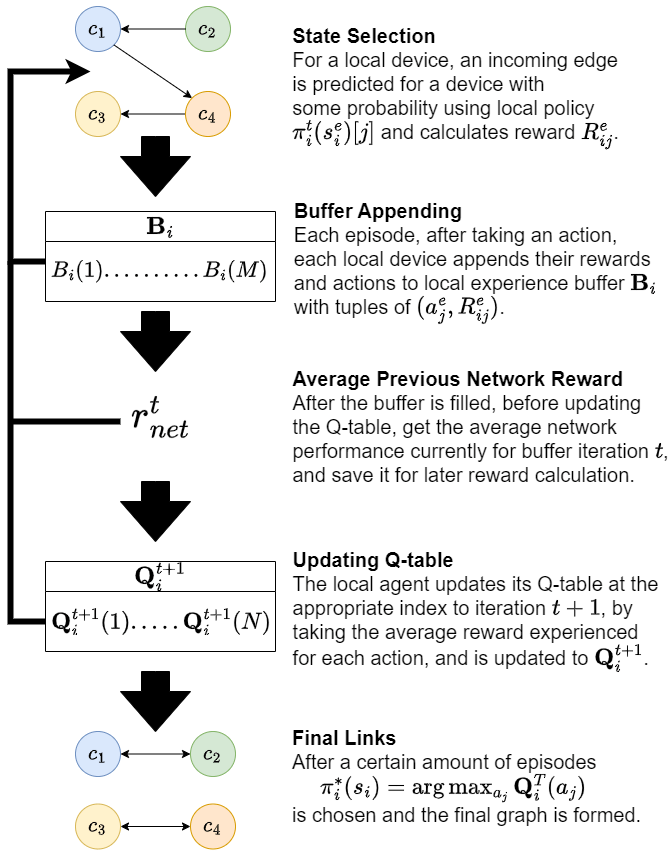} 
    \caption{\small Process of updating the Q-table and selecting the optimal action for a given local device $c_i$}
    \vspace{-0.15 in} 
\end{figure}

To optimize the system-wide metrics, we consider a global reward that includes both $r_{ij}$ and network performance. To that end, we allow devices to share their local rewards between them, which allows each device to calculate the global reward 
\begin{equation}
\begin{aligned}
    R_{ij}^{e} = r_{ij} + \gamma \left(\frac{1}{N}\sum_{i' \in C} r_{i'j'} - r_{net}^{t-1} \right), \label{eq:4}
\end{aligned}
\end{equation}
where devices $i, i'$ select receivers $j,j'$ respectively and $r_{net}^{t-1}$ is the average network performance from the previous update of experience buffer $\textbf{B}_i$, and $\gamma$ is an importance parameter that increases as $t$ does, to place more importance on the network performance as it takes more meaningful actions.
Local devices are designed to decide the best link for data transfer by considering the dissimilarity, local transmission constraints, and system-wide performance when making decisions. To that end, each local device has Q-table $\mathbf{Q}_i \in \mathbb{R}^{T \times N}$, where $\mathbf{Q}_i^t$ corresponds to the entry at time $t$, and is used to accumulate rewards for taking action $a_j$. The table is initialized with small equal values, with agents learning the table by trial and error, allowing for exploration. It first normalizes its Q-table values and predicts an incoming edge $j$ with probability
\begin{equation}
    \pi_i^{t}(s_i^{e})[j] = \frac{\gamma \cdot \frac{\mathbf{Q}_i^{t}(a_j)}{\sum_{k \in C} \mathbf{Q}_i^{t}(a_k)} + \left(1- \gamma \right) \cdot U}{\sum_{k' \in C} \left(\gamma \cdot \frac{\mathbf{Q}_i^{t}(a_{k'})}{\sum_{k \in C} \mathbf{Q}_i^{t}(a_k)} + \left(1- \gamma \right) \cdot U\right)}, \label{eq:5}
\end{equation}
where local policies $\pi_i$ is in set $\{ \pi_i \in \mathbb{R}^{N} \}_{i \in [1,N]}$, $U$ is a value sampled from a uniform distribution in the interval $[0,1]$, and $j \sim \pi_i^{t}(s_i^{e})$ represents the state in the Markov Decision Process (MDP). As $\gamma$ increases, more emphasis is put on the accumulated Q values over $U$.

We now define experience buffer $\textbf{B}_i$, where the rewards for each episode are added, and hold tuples of $(a_j^{e}, R_{ij}^{e})$. In addition, we define $\hat{r}_{k}^{f}$ as the local reward of the link with the highest frequency when the buffer is full for device $k$, i.e.,  $\hat{r}_k^f = \text{arg}\max_{r_{kj}} \sum_{y=1}^{M} \mathds{1}_{\mathbf{B}_k(y)[1] = a_j}$. We use this to get average network performance for iteration $t$ as
\begin{equation}
    r_{net}^{t} = \frac{1}{N} \sum_{k \in C} \hat{r}_{k}^{f}, \label{eq:6}
\end{equation}
which is then used in Equation (\ref{eq:4}) for future rewards. After calculating $r_{net}^{t}$, the Q-table can now be updated as
\begin{equation}
    \mathbf{Q}_{i}^{t+1}(a_j) = \mathbf{Q}_{i}^{t}(a_j) + \frac{\sum_{y=1}^{M} \left( \mathbf{B}_i(y)[2] \cdot \mathds{1}_{\mathbf{B}_i(y)[1] = a_j} \right)} {\sum_{x=1}^{M} \mathds{1}_{\mathbf{B}_i(x)[1] = a_j}} \label{eq:7}
\end{equation}
where it takes the average reward of items that match the corresponding action. $\mathbf{B}_i$ is then cleared for future exploration.

After training, the final links are formed by using the optimal policy, which we define as the action with the maximum accumulated reward, i.e.,

\begin{equation}
    \pi_i^{*}(s_i) = \text{arg}\max_{a_j} \mathbf{Q}_i^{T}(a_j). \label{eq:8}
\end{equation}
Once link formation is finalized, data exchange over the found links can occur, which we explore in the following section.

\subsection{Data Transfers}
After link formation, a decision must be made on which data points to exchange. Anomaly detection \cite{11} has been shown to be effective on autoencoders with data it wasn't trained on. We take advantage of this to find which datapoints in the transmitter will benefit the receiver. As a pre-training step, each device does one iteration of full-batch gradient descent on its data, keeping track of its reconstruction loss, which we define as the mean-squared error of the output of the decoder and input data. With datasets \(\mathcal{D} =  \bigcup_{c_i \in C} \mathcal{D}_{i}\), input data \(d\), output decoded data \(\ell\), and time-step \(\tau\), we have the loss function 
$\mathcal{L}(\phi_{i}^{\tau}, \mathcal{D}_{i}) = \sum_{(d, \ell) \in \mathcal{D}_{i}} \textrm{MSELoss}(\phi_{i}^{\tau}, d, \ell)$, 
where we sum over the reconstruction loss of the mean-squared error. Given this error, we take a subset $\mathcal{K}_{reserve}^{jk}$ from $c_{j_{k}}$, cluster $k$, given that it is permitted by $\mathbf{T}_j[i,k]$. For datapoints from each cluster in $\mathcal{K}_{reserve}^{jk}$, we run it through the receiver autoencoder. If the error is worse than the baseline 
after considering dataset size differences, i.e., $ \frac{\mathcal{L}(\phi_{i}^{\tau}, \mathcal{D}_{i})}{|\mathcal{D}_{i}|} < \frac{\mathcal{L}(\phi_{i}^{\tau}, \mathcal{K}_{reserve}^{jk})}{|\mathcal{K}_{reserve}^{jk}|}$, then we can engage in D2D, due to the indication that there are datapoints that the autoencoder could benefit from in the learning process. 
\\ \\
\noindent 
\textbf{Remark 1:} After D2D exchanges have taken place, we expect that the number of suspected classes in each device should increase, which can mitigate the effect of stragglers. We explore this with numerical simulations in Sec. \ref{sec:simulations}.
\section{Proposed Methodology}

\subsection{Distance and Graph Formation}
We define a graph $\mathcal{G}$ over devices such that there is an edge between two devices if they share a direct communication link. As explained in Section \ref{sec:reward}, we use $K$-means++ on decomposed data $X_i$ to get our clustering assignments and centroids $\mathbf{v}_i$ for each local device, which now can be utilized in the reward accumulation process. 

Firstly, each local device will act as its own local agent, and will be initialized with a Q-table of small equal values. It will then iteratively find links using the Q-learning method.

Each episode during the RL training process, a device selects a transmitter device with some probability \( \pi_i^{t}(s_i^{e})[j]\) from (\ref{eq:5}), and accumulate the rewards in local experience buffer \(\mathbf{B}_i\) with the action \(a_j^{e}\) and global reward \(R_{ij}^{e}\) for episode $e$. If the experience buffer is full at iteration $t$, it is used to update the Q-table to iteration $t+1$ using Equation (\ref{eq:7}). The experience buffer is then cleared. This process continues until the last episode. We then finalize the graph by taking the maximum of the accumulated rewards. After the formation, D2D exchanges can take place between the local devices before the FL process.

\begin{algorithm}[t]
\caption{: Graph Creation with RL.}
\begin{algorithmic}[1]
\State \textbf{Input:} Clients \(c_i \in C\), lower-dimensional PCA data $X_i$, trust matrix $\mathbf{T}_j$, probability of failure $\mathbf{P}_D$.
\State $c_i$ performs K-means++ on \(X_i\) to obtain \(k\) clusters.
\State Initialize empty graph \(\mathcal{G}\), nodes \(\in C\), and \(c_i\)'s Q-table $\mathbf{Q}_i$.
\State $c_i$ predicts edge with probability $ \pi_i^{t}(s_i^{e})[j]$ from (\ref{eq:5}) and computes its local reward $r_{ij}$ from (\ref{eq:3}).
\State $c_i$ computes global reward $R_{ij}^{e}$ from (\ref{eq:4}) for episode $e$ 
\State $c_i$ appends to buffer $\mathbf{B}_i$, gets network performance $r_{net}^{t}$ from (\ref{eq:6}), and updates $\mathbf{Q}_i^{t}$ according to (\ref{eq:7}) when full.
\State $c_i \in C$ forms links based off optimal policy $ \pi_i^{*}(s_i)$ in (\ref{eq:8}).
\State \textbf{Output: } Graph $\mathcal{G}'$ with nodes $\mathcal{V}$ and optimal edges $\mathcal{E}'$
\end{algorithmic}
\end{algorithm}

\subsection{Data Exchange}

The selected links in the graph represent data exchange between clients. For each link formed, we transfer a certain amount of data points. In addition, we also want to make sure the sender only sends datapoints permitted from trust matrix $\mathbf{T}_j$. Afterwards, we find a subset of datapoints $\mathcal{D}_j' \in \mathcal{D}_j$, which is $\bigcup \mathcal{K}_{reserve}^{jk} \in \mathcal{K}_{reserve}^{j}$, and find those that have a worse reconstruction error compared to the baseline.


\begin{algorithm}[t]
\caption{Proposed D2D-Enabled Unsupervised FL}
\begin{algorithmic}[1]
\State \textbf{Input:} Local and global models $\phi_{i}^{\tau}$ and $\phi_{G}^{\tau}$, Clients \(c_i \in C\), optimal finalized edges $\mathcal{E}'$.
\State $\phi_{i}^{\tau}$ does gradient descent on dataset, gets error $\mathcal{L}(\phi_{i}^{\tau}, \mathcal{D}_{i})$.
\State $c_i$ checks if error is worse than reserve points from $c_j$ through one iteration of SGD from (\ref{eq:10}).
\State Each local model trains for $\tau_{a}$ iterations
\State $\phi_{G}^{\tau}$ does global aggregation with a traditional FL algorithm, sends information back for further training.
\State \textbf{Output: } FL with expected enhanced convergence speed.
\end{algorithmic}
\end{algorithm}

\setlength{\textfloatsep}{10pt}%

\subsection{Learning Task}
After D2D transfers have been conducted, we use autoencoders as an unsupervised learning technique to minimize the global loss. Each local device has its own local model, which has the same architecture as the central node. The autoencoder is trained to reconstruct the inputted data to the best of its ability. Our goal is to minimize the global reconstruction loss during the training process, hence, \(\phi_{G}^{*} = \text{arg}\min_{\phi \in \mathbb{R}^{p}} \sum_{i=1}^{N} \mathcal{L}(\phi, \mathcal{D}_{i})\). To that end, local models undergo stochastic gradient descent (SGD), where one iteration is
\begin{equation}
    \phi_{i}^{\tau + 1} = \phi_{i}^{\tau} - \eta \nabla_{\phi_{i}}\frac{\mathcal{L}(\phi_{i}^{\tau}, \mathcal{D}_{i}^{\mathcal{B}}    )}{|  \mathcal{D}_{i}^{\mathcal{B}}  |}, \label{eq:10}
\end{equation}
and $\mathcal{D}_{i}^{\mathcal{B}}$ is a minibatch of $\mathcal{D}_i$ and $\eta$ is the learning rate.
\section{Numerical Analysis}
\label{sec:simulations}

\textbf{\textit{Simulation Setup: }}We conduct simulations on a FL setting having 30 clients on CIFAR-10 \cite{19} and FashionMNIST (FMNIST) \cite{13} datasets to evaluate the performance of the proposed method against the following baselines: (i) no data exchanges (non-i.i.d.) and (ii) graphs generated at uniform.  We run RL for 600 episodes and make the size of experience buffer $M = 90$. We adopt a Convolutional Neural Network (CNN) for both FMNIST and CIFAR-10, and train for 1500 minibatch iterations with aggregation interval $\tau_a = 10$. Each device starts off with 3 classes in a non-i.i.d. distribution. 
We use PyTorch for neural network design and training, and scikit-learn for PCA and $K$-means++, conducted on a desktop with 40GB NVIDIA A100-PCIE GPU and 128GB RAM. 

\textbf{\textit{Heatmaps of Dissimilarity: }}Fig. 3 corresponds to the heatmaps of the dissimilarity between local devices ($\lambda_{ij}$) for FMNIST before and after D2D. In this setup, we use 10 local devices, with $c_i$'s domain of labels being $\{i-1,i,i+1\}$ in a circular fashion. The heatmaps show that when the labels are different, $\lambda_{ij}$ becomes a higher value, indicating dissimilarity between local devices. Furthermore, after D2D, $\lambda_{ij}$ decreases, showing that local data have greater similarity between each other. Our method is shown to effectively find dissimilarity between devices utilizing the centroids from $K$-means++ while also preserving the privacy of the device's data.

\begin{figure}[t!] 
    \centering

        \includegraphics[width=\linewidth]{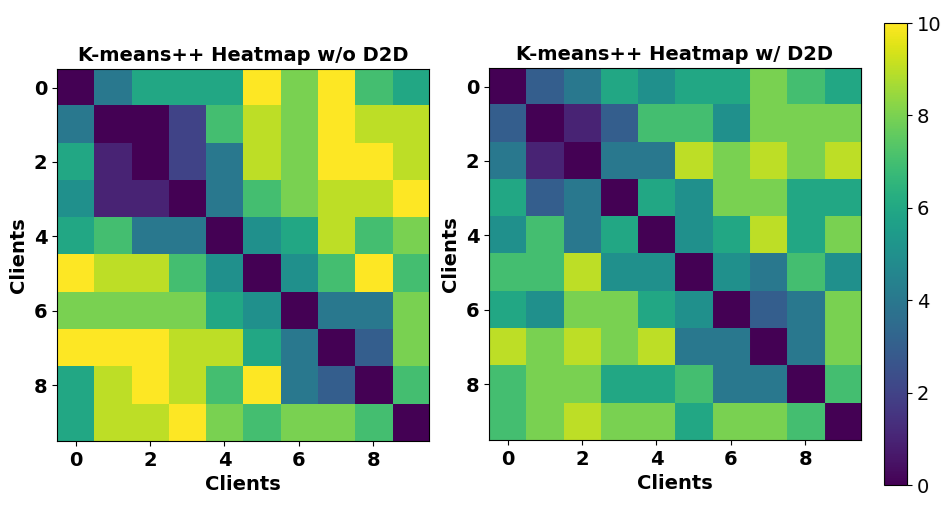}
        \caption{\small Dissimilarity between clients across 10 devices with FMNIST. Left is before D2D and right is after D2D, with average value of 6.24 and 5.61, respectively.}
        \vspace{-0.15 in} 
\end{figure}

\textbf{\textit{Communication Links: }}Fig. 4 corresponds to the probability of failed transmission ($\mathbf{P}_D(i,j)$) of links formed in the graph at uniform versus RL. Doing RL allowed the devices to consistently picks links with low failure, while uniform selection resulted in many devices picking more expensive links, showing that RL finds links effectively based off both dissimilarity and communication cost considerations, indicating robustness to failed transmission within the network.

    \begin{figure}[t!]
        \includegraphics[width=\linewidth, height=38mm]{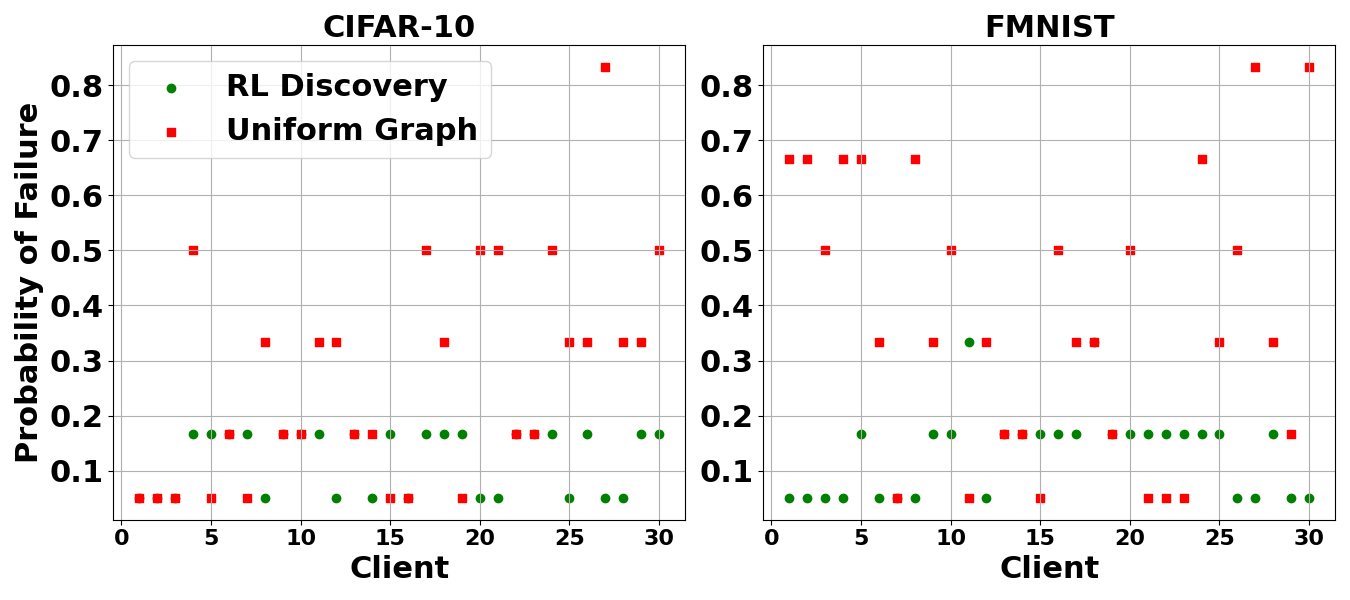}
       \caption{\small Probability of failure of links formed at RL and uniform. Our method significantly improves the probability of successful transmission compared to uniform across both datasets.}
\vspace{-0.177 cm} 
\end{figure}

\textbf{\textit{Varying FL Schemes: }}
We apply our methodology to three FL schemes: FedAvg, FedSGD \cite{1} and FedProx \cite{12} in Fig. 5 for $N=30$ clients. For all three schemes, engaging in RL and smart D2D significantly outperforms the baselines of non-i.i.d. and uniform. We notice that there is also a faster convergence speed among both datasets, showing that D2D transfer has led to the global model better reconstructing the data in a shorter amount of iterations compared to no transfer. For FedAvg, model averaging is benefited with our method as each local device's parameters are less biased towards its own local dataset, improving convergence speed compared to the baselines. FedSGD, which averages the gradients, likewise benefited from having less biased gradients aggregated at the server, allowing for better performance. FedProx's proximity term reduces bias more effectively, as it encourages clients to learn similar representations of the data, even when their local datasets are different. This complements D2D exchange, which allows clients to share their local data with each other, further reducing bias. While engaging in data exchange in uniform does provide some benefit, explicitly targeting the desired datapoints allowed for significant performance increases. 

\textbf{\textit{Linear Evaluation: }} To evaluate the performance of the latent spaces from the embedding vector of the encoder, we adopt linear evaluation \cite{14}, and use $\phi_{G_{Z}}^{\tau}$, which is the server model with only the encoder, $\forall \tau$, to train a linear layer in a supervised manner on top of $\phi_{G_{Z}}^{\tau}$ to perform classification at the server. The linear layer is trained via 1500 and 1000 iterations for CIFAR-10 and FMNIST respectively. While the work focused on linear classification with contrastive learning (CL), the latent embeddings of both autoencoders and CL will separate out the differing classes better the more generalized it is. When using the encoder in a downstream classification task with linear evaluation in Fig. 5, it outperforms both uniform and non-i.i.d. scenarios across FedAvg, FedSGD, and FedProx. This shows that having less biased local models across both FMNIST and CIFAR-10 has allowed the encoder to separate out the latent embeddings better, allowing for greater accuracy and faster convergence.

\begin{figure*}[t!]
    \begin{center}
        \begin{subfigure}{0.93\textwidth}
        \includegraphics[width=0.99\linewidth]{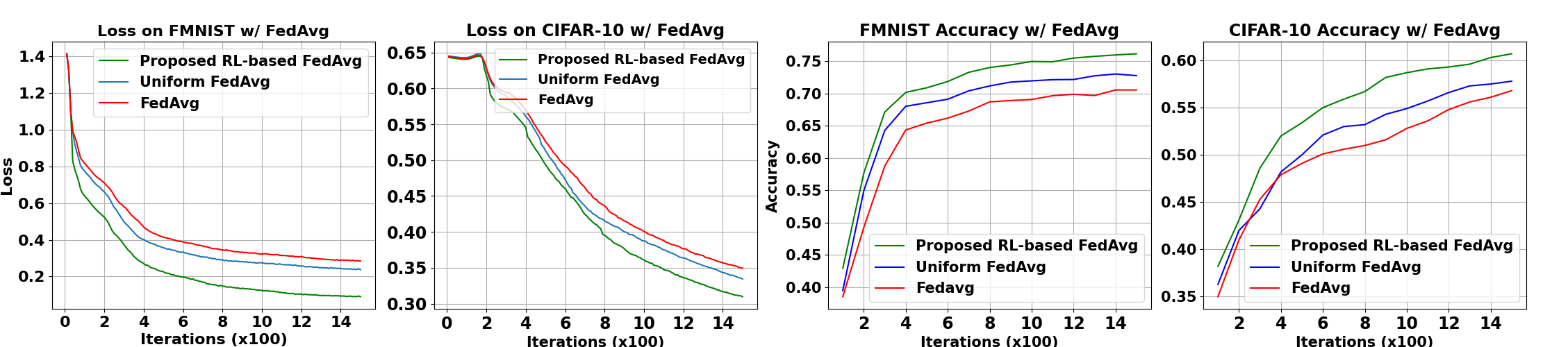}
        \caption{\small}
    \end{subfigure}
    \begin{subfigure}{0.93\textwidth}
        \includegraphics[width=0.99\linewidth]{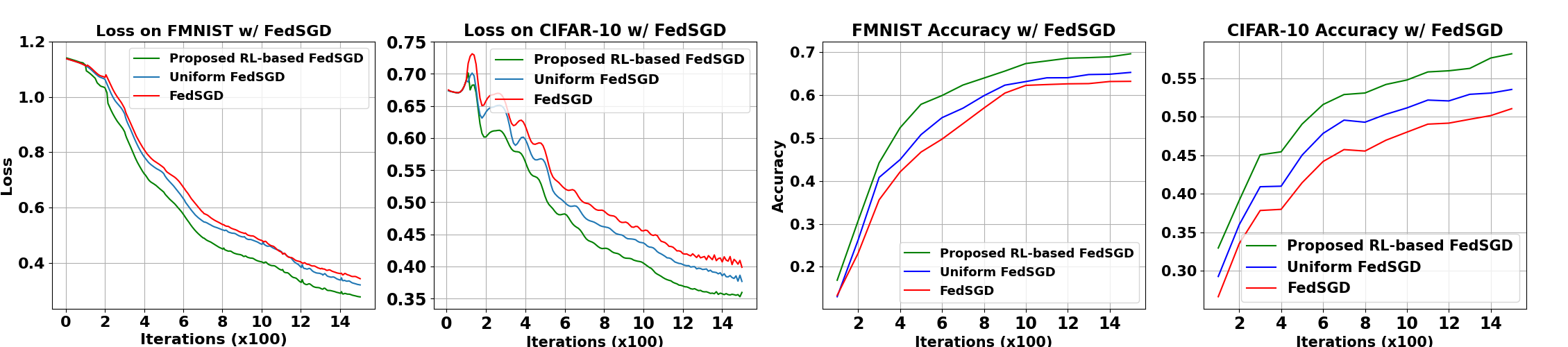}
        \caption{\small}
    \end{subfigure}
    \begin{subfigure}{0.93\textwidth}
        \includegraphics[width=0.99\linewidth]{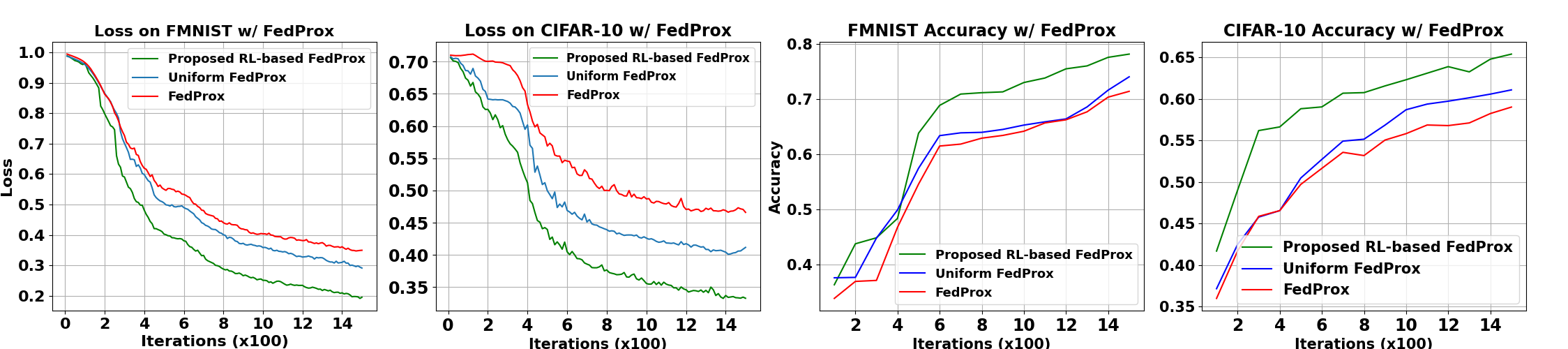}
        \caption{\small}
    \end{subfigure}
    \end{center}
    \vspace{-0.14 in} 
    \caption{\small Reconstruction Loss and Linear Evaluation with (a) FedAvg, (b) FedSGD, and (c) FedProx. Our method significantly improves the loss of the global model across all three FL algorithms carried out on a FL setting with $N=30$ clients. Significant increase in accuracy values also indicates the latent embeddings of the different classes with our proposed method are more spread out, allowing for better classification.}
    \vspace{-0.24 in} 
\end{figure*}

\indent \textbf{\textit{Presence of Stragglers: }} We evaluate the performance of the network with the presence of stragglers, which are excluded from global aggregation. The results of Fig. 6 showcases the robustness of our proposed method compared to the baselines. When less models participate in the aggregation, the model becomes skewed toward their own local non-i.i.d. datasets, but engaging in smart D2D exchange allows the global model to better make up for the inherent bias the stragglers present.

\begin{figure}[t!] 
    \centering
        \includegraphics[width=0.76\linewidth] {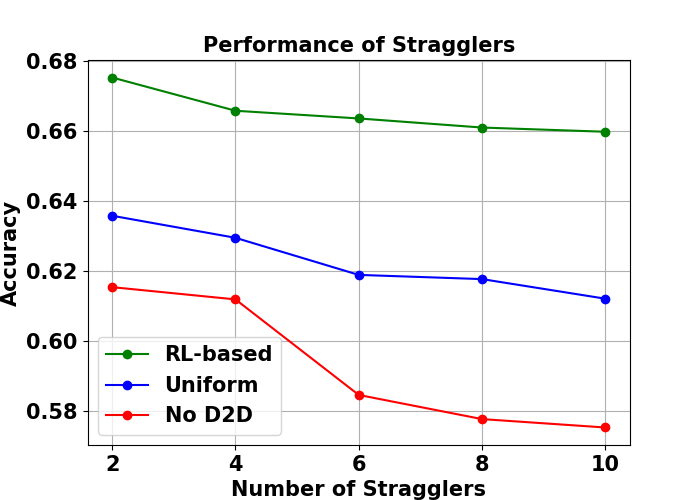}
        \caption{\small Performance of network with different number of stragglers in a setting of $30$ clients. As the number of failures increases, the proposed method is significantly more robust than other approaches.}
        \vspace{-0.22 cm} 
\end{figure}

\vspace{-0.05 cm}
\section{Conclusion}
\vspace{-0.05 cm}
Our numerical analysis showed that the proposed method was robust to different dataset characteristics and communication considerations, such as class imbalance, non-i.i.d. distributions, and failed transmission of data across devices. By leveraging PCA and $K$-means++ in our reward modeling, we successfully created a graph for data transfer that improved the convergence speed, accuracy of downstream tasks, and robustness to stragglers compared to the baselines. Our method is shown to be applicable to multiple FL algorithms, making it a plug-and-chug algorithm that can be placed on top of an unsupervised FL environment. A possible future work may consider a scenario where the overall D2D network involves unstable communication links and is dynamic in nature.

\end{document}